\documentclass[letterpaper]{article}
\usepackage[preprint]{aaai2027}
\usepackage[hyphens]{url}
\usepackage{graphicx}
\usepackage{natbib}
\usepackage{caption}
\usepackage{booktabs}
\usepackage{array}
\usepackage{amsmath}
\usepackage{amssymb}
\newcolumntype{L}[1]{>{\raggedright\arraybackslash}p{#1}}
\title{HERALD: Counterfactual Audits and Minimal Repairs for\\
Proof-of-Retrieval Rewards}
\author{
Zhuowen Liu\textsuperscript{\rm 1}\equalcontrib\corresponding,
Bohan Cui\textsuperscript{\rm 2}\equalcontrib,
YinShang Guo\textsuperscript{\rm 3},\\
Yuting Wang\textsuperscript{\rm 4},
Hao Li\textsuperscript{\rm 5}
}
\affiliations{
\textsuperscript{\rm 1}The Chinese University of Hong Kong\\
\textsuperscript{\rm 2}Fudan University\\
\textsuperscript{\rm 3}Nanjing University\\
\textsuperscript{\rm 4}Institute of Software, Chinese Academy of Sciences\\
\textsuperscript{\rm 5}The Hong Kong University of Science and Technology
}

\begin{document}
\maketitle
\newcommand{\VtwoMainIsolatedN}{593}
\newcommand{\VtwoRzeroLaunderFirst}{4.30\%}
\newcommand{\VtwoRzeroLaunderWorst}{6.66\%}
\newcommand{\VtwoLonlyLaunderWorst}{0.00\%}
\newcommand{\VtwoLonlyZeroUcb}{0.50\%}
\newcommand{\VtwoFullLaunderWorst}{3.37\%}
\newcommand{\VtwoCrossModelN}{3829}
\newcommand{\VtwoCrossModelRzeroWorst}{13.69\%}
\newcommand{\VtwoCrossModelLonlyWorst}{0.00\%}
\newcommand{\VtwoFullCompatibleRetention}{74.45\%}
\newcommand{\VtwoFullCompatibleRetentionCi}{[71.08\%, 77.99\%]}
\newcommand{\VtwoRulfMusiqueEmDelta}{-7.81}
\newcommand{\VtwoRulfMusiqueEmCi}{[-15.62,+0.00]}
\newcommand{\VtwoRvisHotpotEmDelta}{+6.25}
\newcommand{\VtwoRvisHotpotEmCi}{[+1.56,+12.50]}
\newcommand{\VtwoRvisHotpotNaturalDelta}{-4.69}
\newcommand{\VtwoRvisTwoWikiNaturalDelta}{-6.25}
\newcommand{\VtwoRvisMusiqueNaturalDelta}{-6.25}
\newcommand{\VtwoRvisMusiqueRepeatDelta}{-6.25}
\newcommand{\VtwoRvisMusiqueRepeatCi}{[-12.50,-1.56]}

\begin{abstract}
Search-agent rewards mix answer quality, citation grounding, tool cost, and
anti-hacking terms; a high score therefore need not imply that cited evidence was
retrieved, and added penalties can cancel. We introduce \textbf{HERALD}, an offline
audit that applies exact same-question interventions, separates candidate-visible from
oracle information, and enumerates detector contracts before policy optimization. On
four Qwen3-8B pools from HotpotQA, 2WikiMultiHopQA, and MuSiQue, $R_0$ rejects search
deletion and fake IDs, but a label-free citation-laundering attack succeeds. A complete
$2^3$ ablation identifies targeted strengthening of $L$---citing a corpus passage absent
from the retrieved evidence---as the observed inclusion-minimal repair: $R[L]$ has zero empirical ASR with a
\VtwoLonlyZeroUcb{} one-sided cluster upper bound. The gap persists across pool rules,
a visible BM25 attacker, and four models; broader hardening remains vulnerable
when the attack removes an oracle support-ID penalty. Under strict 5M-token matched
training evaluated on 256 paired questions per benchmark, $R[L]$ meets the EM
non-inferiority gate on HotpotQA and 2Wiki but not MuSiQue. Equal-suite citation precision and support recall
improve by 2.02 and 1.46 points, unsupported citations fall by 1.69, and laundering
attackability falls on 2Wiki and MuSiQue. Natural $L$ is not reduced, and the detector
appears in only 18 of 58,368 training trajectories. HERALD thus separates robust
scoring, sparse learning signal, and policy transfer.
\end{abstract}

\section{Introduction}
Reinforcement learning increasingly trains language models to search, call tools, and
answer with citations \cite{searchr1,r1searcher,research,torl,carr,pou}. The reward is
usually decomposed into answer correctness, evidence support, citation format, tool
cost, and action validity. This is convenient because the score can be replayed offline.
It also creates substitution opportunities: answer reward can compensate for a broken
retrieval trace, tool-cost savings can offset an anti-hacking term, and an annotation
proxy can penalize a reasonable original more than an edited attack.

Most work evaluates the policy after optimizing such a reward. We ask a prior question:
\emph{does the reward itself prefer an exact contract violation, and which detector is
actually needed to reverse that preference?} Three design mistakes make this harder
than it appears. First, an edit that changes search, answer, and citations at once does
not isolate proof of retrieval. Second, reporting zero attacks without an eligibility
denominator or a nonzero upper bound overstates certainty. Third, comparing a base reward
only with a large detector bundle cannot reveal whether one check suffices or whether
the bundle creates penalty cancellation.

\begin{figure*}[t]
\centering
\includegraphics[width=\textwidth]{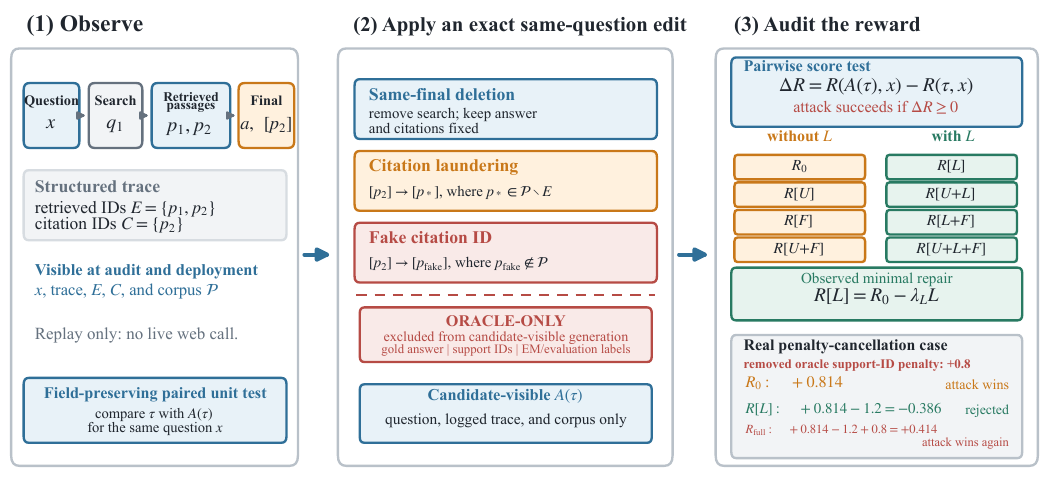}
\caption{HERALD audits a reward before policy optimization. It replays a
structured trajectory, applies field-preserving same-question edits, excludes
gold answers, annotated support IDs, and evaluation labels from candidate-visible
generation. ``Candidate-visible'' qualifies $A$, not $R$. HERALD measures
$\Delta R=R(A(\tau),x)-R(\tau,x)$, with attack success at $\Delta R\geq0$.
Here $L$ flags a corpus-valid citation absent from retrieved passages $E$.
The saved cancellation case shows why the observed inclusion-minimal repair $R[L]$ can
reject an attack that a larger oracle detector bundle prefers again.}
\label{fig:herald_overview}
\end{figure*}

\paragraph{HERALD.}
HERALD treats a reward as a program under paired adversarial unit test. The name HERALD
reflects its role as an early-warning audit that exposes reward loopholes before policy
optimization. It (i) constructs same-question structural interventions; (ii) records
exactly which fields and labels each operator can access; (iii) conditions on attack
eligibility and detector isolation; (iv) reports both attack success and paired reward
margin; and (v) enumerates the full detector lattice. Candidate generation reads only
the question, logged trajectory, and corpus. Candidate-visible refers only to generation;
reward-maximizing selection is an oracle-selected offline upper bound whenever the score
reads gold-derived terms. Figure~\ref{fig:herald_overview} makes this boundary explicit.

\paragraph{Main findings.}
The audit first disproves an over-broad story. The implemented $R_0$ already rejects
same-final search deletion and nonexistent citation IDs. The unresolved loophole is
adaptive \emph{citation laundering}: replacing a citation by a real corpus passage that
the agent never retrieved. A label-free lexical generator exposes this gap under
$R_0$. Exhaustive ablation then shows that the single exact membership check $L$ closes
all three observed primary attacks in the context of $R_0$; the other two exact checks
are redundant on these pools. The failure transfers to four models, while a
broader strict reward retains attacks because its oracle support-ID penalty can be
removed. These are observed, conditional findings, not universal guarantees.

\paragraph{Policy transfer.}
Strict 5M-token, batch-matched Search-GRPO runs differ only by
$R[L]=R_0-\lambda_L L$. On 256 paired questions per benchmark, $R[L]$ improves evidence
quality and targeted attackability on two suites, but not natural $L$ incidence or
uniform EM non-inferiority. Only 18 of 58,368 training trajectories expose the detector,
so the policy result supports transfer under a sparse treatment, not lower natural
violation incidence.

\paragraph{Contributions.}
Our contributions are:
\begin{itemize}
  \item a field-preserving counterfactual audit with explicit information boundaries,
  eligibility, paired margins, and upper bounds for zero events;
  \item a complete $2^3$ detector ablation, cancellation decomposition, and finite-set
  condition identifying the observed inclusion-minimal repair;
  \item label-free lexical/BM25 attacks, pool sensitivity, and four-model replication
  over 3,829 isolated trajectories; and
  \item a strict matched-token policy comparison with a 256-question paired tail
  evaluation, training-signal audit, and reproducible artifacts.
\end{itemize}

\section{Related Work}
\paragraph{Search-agent reinforcement learning.}
RAG, ReAct, and IRCoT established retrieval and interleaved reasoning/action
\cite{rag,react,ircot}. Search-R1, R1-Searcher, ReSearch, ToRL, and DynaSearcher train such
behavior with reinforcement learning \cite{searchr1,r1searcher,research,torl,dyna}; CaRR uses
citation-aware rubrics \cite{carr}, and STAMP assigns credit to the first action exposing
support \cite{stamp}. These methods optimize policies; HERALD first tests a fixed
reward's ordering of controlled violations.

\paragraph{Attribution and proof of use.}
Attribution work tests or repairs whether claims are supported by identified evidence
\cite{gophercite,ais,alce,rarr,autoattrib,verifiability}; Proof-of-Use links retrieval,
reasoning, and the answer \cite{pou}. Proof of retrieval is narrower: a cited ID must
occur in prior observations. This candidate-visible check is inexpensive but does not
establish entailment, so HERALD separates it from correctness and support-ID agreement.

\paragraph{Counterfactual auditing.}
CheckList and contrast sets motivate controlled edits \cite{checklist,contrastsets};
RATE estimates reward sensitivity under imperfect rewrites \cite{rate}, while
Counterfactual Trace Auditing pairs traces with and without a skill \cite{cta}. HERALD
targets executable contracts with exact edits, detector-specific eligibility, and
detector-lattice decisions.

\paragraph{Reward hacking and verifier reliability.}
Reward misspecification, corrupted channels, causal tampering, and reward gaming explain
proxy failure
\cite{concrete,corruptedreward,rewardtampering,rewardgaming,rewardmisspec};
reward-model overoptimization and RewardBench expose learned-proxy limits
\cite{rewardoveropt,rewardbench}. Recent reward-hacking benchmarks, judge attacks,
spurious RLVR rewards, and imperfect-verifier analyses show the same risk in LMs
\cite{rhb,onetoken,spurious,gaming,noisy,imperfect,sober}. HERALD adds an exact negative
control and subset ablation to falsify the loophole or need for a larger defense.

\section{Problem Setup}
\paragraph{Search trajectories.}
For question $x$, a trajectory $\tau$ contains ordered search actions, retrieved
observations, a final answer, and citation IDs. Let $E(\tau)$ be the set of passage IDs
actually returned to the agent, $C(\tau)$ the final citation set, and $\mathcal P$ the
fixed corpus-ID set. The audit has access to the structured log and corpus index; no live
web calls are made.

\paragraph{Base reward.}
The existing $R_0$ combines answer quality $a$, grounding $g$, output constraints $c$,
tool cost $t$, invalid-action count $v$, a generic hacking rate $h$, and length $\ell$:
\begin{equation}
R_0=1.15a+1.05g+0.30c-0.04t-0.80v-0.55h-0.004\ell.
\end{equation}
The generic rate averages seven Boolean flags, including under-retrieval,
cited-not-retrieved, fake citation, support-ID mismatch, and repeated query. Because
$L$ is one of them, an isolated $L$ flip already contributes $-0.55/7\approx-0.0786$
inside $R_0$; $R[L]$ strengthens an existing detector rather than introducing one.
We also audit outcome-only and outcome-plus-cost controls.

\paragraph{Exact visible detectors.}
We study three deterministic operators:
\begin{align}
U(\tau)&=\mathbb{1}\{\text{nonempty final with no search}\},\\
L(\tau)&=\mathbb{1}\{\exists c\in C(\tau)\cap\mathcal P:c\notin E(\tau)\},\\
F(\tau)&=\mathbb{1}\{\exists c\in C(\tau):c\notin\mathcal P\}.
\end{align}
Thus a nonexistent ID belongs only to $F$, never to $L$.
For $S\subseteq\{U,L,F\}$,
\begin{equation}
R_S(\tau,x)=R_0(\tau,x)-\sum_{j\in S}\lambda_j j(\tau),
\end{equation}
The audit manifest fixes $\lambda_U=1.6$, $\lambda_L=1.2$, and $\lambda_F=1.0$ for every
lattice cell and attack; these targeted terms are added on top of the generic $h$ term.
$R_{\mathrm{full}}$ additionally includes answer-without-
citation, repeated-query, over-retrieval, and oracle support-ID penalties. The latter
reads benchmark annotations and is never called candidate-visible.

\section{HERALD Audit}
\paragraph{Exact operators.}
The same-final operator deletes all search actions and observations while preserving the
question, answer, ordered citation IDs, and raw final output. The fake-ID operator
replaces a citation by a nonexistent corpus ID. The adaptive laundering operator replaces
it by a real but unretrieved ID. The primary generator builds up to eight candidates by
lexical overlap with the visible question and candidate answer. A separate BM25 generator
uses the same visible fields and evaluates budgets $K\in\{1,2,4,8,16\}$. Neither receives
an example record, gold answer, support ID, EM label, or evaluation field.

\paragraph{Two candidate policies.}
\emph{First-visible} selects the top candidate using only candidate-visible fields; it is
the label-free attack selector, although its ASR under $R$ remains an offline statistic.
\emph{Oracle-worst} maximizes $R$ over the
same candidates. Because $R_0$, $R[L]$, and $R_{\mathrm{full}}$ read held-out answer or
support terms, this is an oracle-selected upper bound, not label-free selection.
Generation remains label-free in both cases.

\begin{table}[t]
\centering\small
\setlength{\tabcolsep}{2.5pt}
\begin{tabular}{@{}L{0.25\columnwidth}L{0.43\columnwidth}L{0.21\columnwidth}@{}}
\toprule
Operator & Preserved / changed & Generation \\
\midrule
same-final & final bytes fixed; remove search & visible exact \\
laundering & answer/search fixed; replace citation & visible ranked \\
fake ID & answer/search fixed; replace citation & visible exact \\
gold stress & answer/citations become gold & oracle only \\
support-ID & cite non-annotated retrieval & oracle only \\
\bottomrule
\end{tabular}
\caption{Primary interventions and information boundary.}
\label{tab:attacks}
\end{table}

\paragraph{Eligibility and isolation.}
The operator target sets are
$D_{\mathrm{same}}=\{U,L\}$, $D_{\mathrm{launder}}=\{L\}$, and
$D_{\mathrm{fake}}=\{F\}$. An attack is eligible when every detector in $D_A$ changes
from zero to one; it is isolated when every visible detector outside $D_A$ is unchanged.
We report constructible, eligible, and isolated-eligible subsets; primary tables use the
last, preventing a zero caused by attacking already-invalid originals.

\paragraph{Paired estimands.}
For attack $A$, the paired margin and attack-success rate are
\begin{align}
M_R(A,\tau)&=R(A(\tau),x)-R(\tau,x),\\
\mathrm{ASR}_R(A)&=\mathbb{E}\left[\mathbb{1}\{M_R(A,\tau)\ge0\}\right].
\end{align}
Ties count as attacks. Questions, not trajectory rows, are the sampling unit. Duplicate
MuSiQue pools share a normalized-question cluster. We use 5,000 deterministic
question-cluster bootstrap replicates for means and intervals. If $n$ independent
question clusters have zero events, we report the one-sided exact 95\% upper bound
$1-0.05^{1/n}$ rather than a degenerate bootstrap interval.

\paragraph{Complete detector lattice.}
We evaluate all eight subsets of $\{U,L,F\}$. A subset is \emph{observed sufficient} when
all primary Oracle-worst ASRs are zero on isolated-eligible questions. It is
\emph{observed inclusion-minimal} when it is sufficient and no proper subset is
sufficient. Because the full $2^3$ lattice is enumerated, this checks every proper subset
rather than assuming monotonicity. This
definition is explicitly conditional on $R_0$, the attacks, candidate generator, and
observed pools.

\paragraph{Contract-compatible retention.}
An original is contract-compatible when it has a valid nonempty final, one to five search
calls, nonempty known citations all previously retrieved, and no repeated-query pair
above the fixed threshold. Retention is the fraction of this rule-defined subset receiving
no targeted penalty. It is not a human false-positive rate or a semantic-support label.

\section{Experimental Setup}
\paragraph{Main pools.}
The main audit uses four existing Qwen3-8B best-of-eight pools: 200 questions each from
HotpotQA and 2WikiMultiHopQA, and 200 MuSiQue questions each at retrieval cutoffs five and
eight \cite{hotpot,twowiki,musique}. The MuSiQue pools repeat question IDs, so 800 rows
represent 600 unique questions. Exactly 593 question clusters are isolated-eligible for
all three primary attacks.

\paragraph{Cross-model replication.}
We reuse 1,000 logged HotpotQA trajectories for each of Qwen3-1.7B, Qwen3-8B,
Qwen3-14B, and Meta-Llama-3-8B under a fixed corpus, parser, attack implementation, and
reward implementation. After isolation filtering, the per-model sample sizes are
904--995. These are existing trajectories, not new GPU generations.

\paragraph{Sensitivity and utility.}
We multiply the canonical $U/L/F$ penalties by
$\{0,0.25,0.5,1,2\}$. Separately, we rescore the saved best-of-eight candidate pools and
select by each audited reward. Because task reward uses gold answers and support IDs, this
is an offline oracle utility diagnostic, never an inference-time selector.

We also rerun the isolated laundering audit under stored selection, first sample,
question-hash random sample, all eight samples, $R_0$-best, and explicitly
gold-informed best-of-eight. First, hash-random, and all-eight do not maximize a gold
score; $R_0$-best and gold-informed are oracle, while stored selection is inherited.
BM25 candidates are visible; maximizing $R$ over them is Oracle-worst.

\paragraph{Matched policy audit.}
We analyze completed eight-GPU Search-GRPO runs with group size eight and a strict
5,000,000 global generated-action-token budget, stopping at the first complete optimizer
update above the budget. Batch-matched $R_0$ and $R[L]$ both complete at update 912 with
5,003,240 and 5,003,684 tokens. Their base checkpoint, seed, question schedule,
generation, optimizer, advantage construction, and code are identical; only the reward
configuration adds $-\lambda_L L$. All 192 update-0 non-reward records match exactly.
Each arm stores 58,368 training trajectories.

The final tail evaluation uses 256 nontraining questions per benchmark, one greedy
trajectory and four fixed-seed samples per question. Across the shared base, $R_0$, and
$R[L]$ arms, this gives 11,520 trajectories; question IDs, generation seeds, and
generation protocols are exactly paired. Intervals use question-cluster bootstrap.
The pre-specified EM non-inferiority gate requires the paired lower confidence bound to
exceed $-2$ points on each benchmark. A targeted evaluation constructs up to 16
candidate-visible laundering edits per generated trajectory; First-visible is the
label-free selector, while audited-reward maximization is an Oracle-worst upper bound.

\begin{figure*}[t]
\centering
\includegraphics[width=\textwidth]{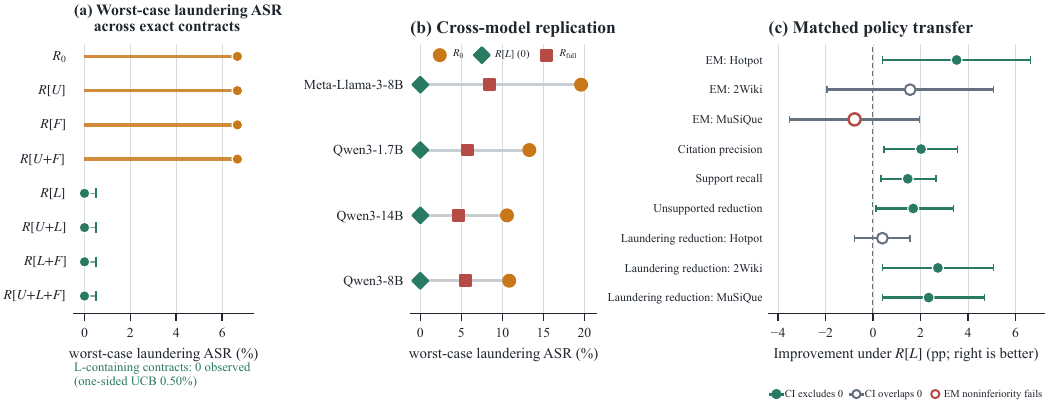}
\caption{Audit-to-policy evidence. (a) Complete $2^3$ exact-contract
ablation for worst-case laundering: every subset lacking $L$ retains 6.66\%
ASR, whereas every $L$-containing subset has zero observed ASR (one-sided
cluster UCB 0.50\%); same-final and fake-ID ASR are already zero under $R_0$.
(b) The $R_0$ failure, $R[L]$ repair, and residual $R_{\mathrm{full}}$ tail
replicate across four models. Panels (a,b) use Oracle-worst selection;
Table~\ref{tab:v2_core_audit} also reports First-visible. (c) Matched 5M-token policy effects on
256 paired greedy questions per benchmark. EM, citation precision, and support
recall use $R[L]-R_0$; unsupported-citation and laundering-ASR reductions use
$R_0-R[L]$, so right is uniformly better. Intervals are paired 95\%
question-bootstrap CIs. Green intervals exclude zero, open gray intervals
overlap zero, and the red ring marks failed $-2$ point EM non-inferiority.}
\label{fig:core_results}
\end{figure*}

\section{Reward-Audit Results}
Table~\ref{tab:v2_core_audit} exposes the headline audit counts and margins directly;
Figure~\ref{fig:core_results}(a) then expands the comparison to the complete detector
lattice.

\begin{table*}[t]
\centering\small
\setlength{\tabcolsep}{5pt}
\begin{tabular}{@{}lrrrrr@{}}
\toprule
Reward & Same-final & Launder First-visible & Oracle-worst & Fake ID & Mean max margin \\
\midrule
$R_0$ & 0 (0.50) & 4.30 & 6.66 & 0 (0.50) & -0.489 \\
$R[L]$ & 0 (0.50) & 0 (0.50) & 0 (0.50) & 0 (0.50) & -1.689 \\
$R[U{+}L{+}F]$ & 0 (0.50) & 0 (0.50) & 0 (0.50) & 0 (0.50) & -1.689 \\
$R_{\mathrm{full}}$ & 0 (0.50) & 1.69 & 3.37 & 0 (0.50) & -2.132 \\
\bottomrule
\end{tabular}
\caption{Paired exact-contract audit on isolated-eligible trajectories. First-visible is
label-free; Oracle-worst maximizes the audited reward over the same eight candidates and is an
offline upper bound. ASR is in percent; zeros have one-sided 95\% exact cluster UCBs.}
\label{tab:v2_core_audit}
\end{table*}

\subsection{Exact controls falsify two broad loopholes}
Outcome-only rewards give ASR 1 for same-final, laundering, fake-ID, repeated-query, and
oracle unsupported-citation attacks. Adding tool cost still gives ASR 1 for all except
repeated query; search deletion is positively rewarded because it saves cost. In contrast,
$R_0$ has zero same-final and fake-ID ASR on 593 isolated questions. The exact upper bound
is \VtwoLonlyZeroUcb{}, so the correct statement is ``no observed event below this
resolution,'' not population risk zero.

For same-final deletion, $R_0$ preserves answer and grounding while removing tool cost
and activating generic under-retrieval and cited-not-retrieved flags. The implemented
generic penalty already exceeds the saved cost. This negative control rules out search
deletion itself as the baseline vulnerability.

\subsection{Adaptive laundering exposes the actual gap}
Citation laundering leaves answer and search fixed but substitutes a real corpus ID absent
from prior observations. Under $R_0$, the First-visible candidate succeeds
\VtwoRzeroLaunderFirst{} and the Oracle-worst candidate succeeds
\VtwoRzeroLaunderWorst{}. The corresponding average
worst-case margin is $-0.489$: most attacks lose, but a nontrivial tail ties or wins.

$R[L]$ reduces both First-visible and Oracle-worst ASR to zero, with mean worst-case
margin $-1.689$. $R[U{+}L{+}F]$ is identical because $U$ and $F$ were already handled
by $R_0$. $R_{\mathrm{full}}$ is stronger on average yet retains
\VtwoFullLaunderWorst{} Oracle-worst ASR: replacing a citation can remove its active
oracle support-ID penalty and partially cancel the new membership penalty.

\paragraph{The gap is not a best-of-eight or lexical artifact.}
Under stored, first, hash-random, all-eight, $R_0$-best, and gold-informed pool
selection, First-visible $R_0$ isolated laundering ASR stays between 3.96\% and 4.68\%;
Oracle-worst ASR stays between 7.08\% and 7.81\%. $R[L]$ has no successful isolated
attack. Table~\ref{tab:bm25_curve} gives an independent visible BM25 ranking: its $K=1$ visible
selector yields 4.38\% offline ASR, while the oracle tail reaches 13.74\% at $K=16$.
Fake-only attacks are unchanged under $R_0$ and $R[L]$, as required by the disjoint
definitions of $L$ and $F$.

\begin{table}[t]
\centering\small
\setlength{\tabcolsep}{3.5pt}
\begin{tabular}{@{}rrrr@{}}
\toprule
$K$ & $R_0$ visible & $R_0$ oracle & $R[L]$ \\
\midrule
1  & 4.38 & 4.38  & 0.00 \\
2  & 4.38 & 6.41  & 0.00 \\
4  & 4.38 & 9.78  & 0.00 \\
8  & 4.38 & 12.48 & 0.00 \\
16 & 4.38 & 13.74 & 0.00 \\
\bottomrule
\end{tabular}
\caption{Visible BM25 laundering sensitivity on 593 isolated-eligible
questions. Candidate generation is label-free. Oracle-selected worst-case maximizes
the audited reward within the same visible candidate set.}
\label{tab:bm25_curve}
\end{table}

\subsection{One strengthening is observed inclusion-minimal}
The complete lattice in Figure~\ref{fig:core_results}(a) makes the minimality
claim testable. $R_0$, $R[U]$, $R[F]$, and $R[U{+}F]$ all retain the same 6.7\%
Oracle-worst laundering failure. Every subset containing $L$ has zero observed primary
ASR. Consequently $R[L]$, not the old three-check bundle, is the observed inclusion-minimal
contract for these attacks in the context of $R_0$.

Weight sensitivity supports a margin, not merely threshold, interpretation. At scale
zero, laundering ASR is 6.7\%; at scales 0.25 and 0.5 it is 4.0\% and 3.2\%; the
canonical scale one reaches zero and shifts mean margin from $-0.489$ to $-1.689$.
Doubling the penalty preserves zero and increases separation. Same-final and fake-ID ASR
remain zero at every scale because the base reward already rejects them.

\subsection{Broader strict hardening is not monotone}
On the independently defined 593-question compatible set, $R[L]$ and
$R[U{+}L{+}F]$ retain every trajectory by construction. $R_{\mathrm{full}}$ retains
only \VtwoFullCompatibleRetention{}
(\VtwoFullCompatibleRetentionCi{}). This is rule-conditioned retention, not human error:
the difference is concentrated in an oracle support-ID proxy that treats benchmark
annotations as exhaustive.

The same cancellation explains why a detector bundle can have a more negative average
margin yet higher ASR than a smaller contract. For pair $(\tau,A(\tau))$,
\begin{equation}
M_S=M_0-\sum_{j\in S}\lambda_j
\left[f_j(A(\tau))-f_j(\tau)\right].
\end{equation}
If the original already triggers $j$ and the attack removes it, the bracket is negative
and the ``penalty'' raises the attack margin. Monotonic nonnegative weights do not imply
monotonic paired robustness on imperfect originals.

\paragraph{A pointwise monotonicity criterion.}
For a fixed eligible pair set, let
$d_j(A,\tau)=f_j(A(\tau))-f_j(\tau)$ and define the cancellation credit added when
expanding $S$ to $T\supseteq S$ as
\begin{equation}
C_{T\setminus S}(A,\tau)
=-\sum_{j\in T\setminus S}\lambda_j d_j(A,\tau).
\end{equation}
The expanded margin is therefore
\begin{equation}
M_T(A,\tau)=M_S(A,\tau)+C_{T\setminus S}(A,\tau).
\end{equation}
If every added detector has $d_j(A,\tau)\geq0$ for every eligible pair, then
$C_{T\setminus S}\leq0$ pointwise, so both margins and ASR are monotone nonincreasing.
Conversely, a pair rejected by $S$ re-enters the attack set under $T$ exactly when
\begin{equation}
C_{T\setminus S}(A,\tau)\geq-M_S(A,\tau).
\end{equation}
This condition needs no distributional assumption. In the saved case,
$d_{\mathrm{support}}=-1$, hence $C=0.8$ and the $R[L]$ margin $-0.38625$
becomes $+0.41375$ under $R_{\mathrm{full}}$.

\paragraph{Finite-set guarantee and real cancellation case.}
For an edit with $L(\tau)=0$ and $L(A(\tau))=1$, additive $R[L]$ strictly rejects every
attack in a finite candidate set whenever
\begin{equation}
\lambda_L>\max_{\tau,A}\left[R_0(A(\tau))-R_0(\tau)\right].
\end{equation}
The audited maximum $R_0$ gain is $0.81375$, below the manifest-fixed
$\lambda_L=1.2$. The same fixed value is used for every lattice cell and attack; this
comparison is a post hoc finite-set certificate, not a population guarantee.
Offline lexicographic selection by $(-L,R_0)$ therefore agrees with
additive $R[L]$ on the observed candidate sets: both select zero lexical or hash-random
nonlexical attacks, whereas $R_0$ selects 7.32\% and 0.84\%.

Figure~\ref{fig:herald_overview} visualizes a saved MuSiQue pair, not a hand-built
example. The answer and search trace are unchanged; the attack replaces the citation
by a real but unretrieved passage. The $1.2$ membership penalty reverses the
$+0.81375$ base margin, but $R_{\mathrm{full}}$ simultaneously removes a $0.8$
oracle support-ID penalty and makes the attack preferable again.
Among originals with an active support-ID penalty, $R_{\mathrm{full}}$ First-visible
and Oracle-worst ASR are 6.33\% and 13.92\%; without it they are 0.00\%
and 0.22\%. Every successful active-stratum pair removes the 0.8 penalty.

\subsection{The laundering gap transfers across models}
Every model exhibits positive $R_0$ adaptive-laundering ASR
(Figure~\ref{fig:core_results}(b)). Oracle-worst ASR ranges from 10.57\% to 19.60\%;
the question-weighted aggregate is \VtwoCrossModelRzeroWorst{} over
\VtwoCrossModelN{} isolated trajectories. $R[L]$ has zero observed success for all four
models, with per-model upper bounds of 0.30--0.33\%. The strict full reward retains
4.70--8.44\% worst-case ASR. Thus the main failure and the smaller repair are not artifacts
of one Qwen3-8B candidate pool.

\paragraph{Offline utility.}
On the saved best-of-eight pools, $R[L]$ and $R[U{+}L{+}F]$ select exactly the same
aggregate EM, citation precision, support recall, valid-final rate, and search cost as
$R_0$. $R_{\mathrm{full}}$ changes citation precision/support recall by only
$+0.002/+0.001$ but reduces valid-final retention. This is an oracle selection diagnostic,
not an inference-time accuracy result.

\section{Policy-Transfer Results}
Table~\ref{tab:strict_l_policy} gives the per-benchmark matched-policy estimates in a
common improvement direction; Figure~\ref{fig:core_results}(c) emphasizes their
uncertainty and the pre-specified non-inferiority decision.

\begin{table*}[t]
\centering\small
\resizebox{\textwidth}{!}{%
\begin{tabular}{@{}lcccccc@{}}
\toprule
Dataset & $\Delta$ EM [95\% CI] &
$\Delta$ citation precision &
Unsupported reduction &
Laundering-ASR reduction &
strict $L$ (\%) $R_0/R[L]$ &
EM NI \\
\midrule
HotpotQA & $+3.52\ [0.39,6.64]$ & $+2.47\ [0.13,4.95]$ &
$0.00\ [-2.34,2.34]$ & $+0.39\ [-0.78,1.56]$ & $0.00/0.39$ & yes \\
2Wiki & $+1.56\ [-1.95,5.08]$ & $+1.43\ [-1.30,4.36]$ &
$+1.95\ [-1.17,5.08]$ & $\mathbf{+2.73\ [0.39,5.08]}$ & $0.00/0.00$ & yes \\
MuSiQue & $-0.78\ [-3.52,1.95]$ & $+2.15\ [-0.78,5.08]$ &
$\mathbf{+3.12\ [0.39,6.25]}$ &
$\mathbf{+2.34\ [0.39,4.69]}$ & $0.00/0.00$ & no \\
\midrule
Equal-suite & $+1.43\ [-0.39,3.26]$ & $\mathbf{+2.02\ [0.46,3.56]}$ &
$\mathbf{+1.69\ [0.13,3.39]}$ & -- & $0.00/0.13$ & 2/3 \\
\bottomrule
\end{tabular}
}
\caption{Expanded matched policy comparison on 256 paired greedy questions
per benchmark. EM and citation precision use $R[L]-R_0$; unsupported-citation
and laundering-ASR reductions use $R_0-R[L]$. Thus positive is uniformly better.
Laundering uses the first candidate-visible edit and evaluates attack success
with the fixed base scorer $R_0$. Equal-suite intervals resample
questions within each benchmark before averaging. EM non-inferiority (NI)
requires the paired lower bound to exceed $-2$ points.}
\label{tab:strict_l_policy}
\end{table*}

\paragraph{Strict $R[L]$ training improves evidence quality under matched compute.}
Figure~\ref{fig:core_results}(c) summarizes the matched endpoint. On 256 greedy
questions per suite, equal-suite EM changes by $+1.43$ points
($[-0.39,3.26]$), while the $-2$ point EM non-inferiority gate passes on HotpotQA and
2Wiki but not MuSiQue.
Citation precision and support recall improve by $+2.02$ ($[0.46,3.56]$) and
$+1.46$ ($[0.33,2.64]$), while unsupported citations fall by $1.69$ points
(95\% CI for $R[L]-R_0$: $[-3.39,-0.13]$). Four-sample evaluation corroborates this: precision improves by
$+1.72$ ($[0.50,2.93]$), recall by $+1.73$ ($[0.90,2.56]$), and unsupported
citations fall by $1.69$ points (CI for $R[L]-R_0$: $[-2.96,-0.42]$), while EM and F1 intervals cross zero.

\paragraph{Targeted attackability improves on two suites.}
For the First-visible laundering edit, attackability under the fixed base scorer $R_0$
falls by $2.73$ points on 2Wiki (95\% CI for $R[L]-R_0$: $[-5.08,-0.39]$)
and $2.34$ on MuSiQue ($[-4.69,-0.39]$). The HotpotQA difference is $-0.39$
($[-1.56,0.78]$). Four-sample and Oracle-worst audits preserve the two-suite pattern; fake-citation
controls remain unchanged.

\begin{figure*}[t]
\centering
\includegraphics[width=\textwidth]{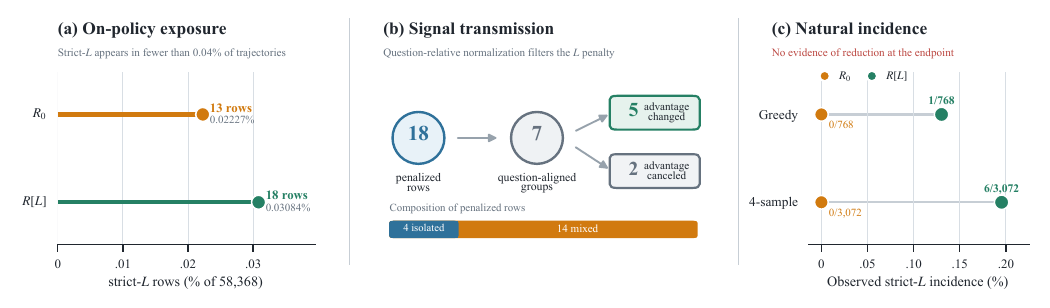}
\caption{The strict-$L$ training signal is active but sparse. $R[L]$ exposes
18 of 58,368 trajectories; only seven question-aligned groups receive the
penalty, and group normalization cancels it in two. At the expanded natural
endpoint, the data provide no evidence of reduced natural strict-$L$ incidence
relative to $R_0$.}
\label{fig:training_signal}
\end{figure*}

\paragraph{The detector reaches the optimizer, but rarely.}
Figure~\ref{fig:training_signal} audits the realized training signal. $R[L]$ sees
18 strict-$L$ rows among 58,368 trajectories (0.03084\%), spanning seven
groups; five change normalized advantages and two are exactly canceled. In the expanded
evaluation, $R_0$ emits no strict-$L$ event among 768 greedy or 3,072 sampled
trajectories; $R[L]$ emits one and six. The implementation therefore transmits a sparse
signal, but natural $L$ is not reduced.

\paragraph{Why group normalization can erase a correct detector.}
Consider one GRPO group of $K$ rollouts with pre-detector rewards $r_i$ and binary
strict-$L$ indicators $\ell_i$. The repaired rewards are
$r'_i=r_i-\lambda_L\ell_i$. Writing group means as $\bar r_g$ and
$\bar\ell_g$, the centered reward obeys
\begin{equation}
\widetilde r'_i
=r'_i-\bar r'_g
=(r_i-\bar r_g)-\lambda_L(\ell_i-\bar\ell_g).
\end{equation}
Thus the detector reaches a relative advantage only through within-group contrast.
For binary $\ell_i$, the squared magnitude of its centered contribution is
\begin{equation}
\frac{1}{K}\sum_{i=1}^{K}
\left(\widetilde r'_i-\widetilde r_i\right)^2
=\lambda_L^2\bar\ell_g(1-\bar\ell_g).
\end{equation}
If every rollout in the group has the same detector value, this quantity is zero:
subtracting the group mean removes the penalty exactly, and subsequent
standard-deviation normalization cannot restore it. A mixed group is therefore
necessary, though not sufficient, for $L$ to change normalized advantages. This
identity explains why row-level exposure overstates optimizer-visible exposure and why
two of the seven penalized groups are exactly canceled.

\section{Implications}
\paragraph{Audit the smallest visible contract.}
Full enumeration shows that only targeted $L$ strengthening adds observed robustness
beyond $R_0$. Justify each bundle component using attack coverage and compatible-set
retention. Candidate-visible operators may read the trace and corpus, but any selection
rule that reads gold answers, support IDs, or evaluation fields must be labeled oracle.

\paragraph{Separate score preference from behavior incidence.}
Counterfactual ASR tests scorer preference on saved pairs; natural incidence requires
fresh rollouts and enough events. The 256-question tail supports better targeted
attackability, not lower natural $L$ incidence. Sparse 0.03\% exposure and
group-normalization cancellation show why both estimands are needed.

\paragraph{Use margins and upper bounds.}
ASR hides severity and vulnerable tails. Report paired margins and an exact upper bound
with $n$: zero observed events do not prove absolute robustness.

\section{Limitations}
The audit tests structured citation IDs and local retrieval; free-form citations,
semantic paraphrases, live-web state, and corpus poisoning need other operators.
Cross-model replication covers only saved HotpotQA trajectories. Lexical, BM25, and
hash-random candidates expose the gap, but dense retrieval and visible-only LLM attacks
remain untested. Oracle-worst, $R_0$-best, and gold-informed selection are labeled
oracle because their scores can read gold-derived terms.

Zero-ASR, minimality, and finite-set margins are conditional on the pools, eligibility
rules, generators, and closed corpus; they do not certify arbitrary future trajectories
or changing indexes.

Proof of retrieval is not proof of semantic support. $R[L]$ can verify that a citation was
observed, but not whether it entails the answer; annotated support IDs are also
incomplete.

The strict policy comparison uses one matched run per arm; paired intervals measure
question uncertainty, not between-run variability. EM non-inferiority holds on two of
three benchmarks, while seven natural $R[L]$ events provide too little power for small
rare-event changes. The data block a reduction claim but do not establish equal rates.

\section{Conclusion}
HERALD makes proof-of-retrieval rewards falsifiable through paired audits. Exact controls
isolate citation laundering and identify targeted $L$ strengthening as the observed
inclusion-minimal repair; cancellation explains why broader oracle hardening can be
weaker. Pool sensitivity, BM25 and nonlexical candidates, and four-model replication
support the scoring result. Matched-token $R[L]$ training improves aggregate evidence
quality and targeted attackability on two benchmarks, but not natural $L$ incidence or
uniform accuracy non-inferiority. Reward audits should separate scorer robustness,
training exposure, and policy transfer.

\clearpage
\bibliography{main}

\end{document}